\documentclass{optica-article}

\journal{opticajournal} % for journals or Optica Open

\articletype{Research Article}

\usepackage{lineno}
\usepackage{times}
\usepackage{epsfig}
\usepackage{graphicx}
\usepackage{amsmath}
\usepackage{booktabs}

\usepackage{multirow}
\usepackage{bbding}
\usepackage{adjustbox}
\usepackage{makecell}

\begin{document}

\title{E-3DGS: 3D Gaussian Splatting with Exposure and Motion Events}

\author{Xiaoting Yin,\authormark{1,†} Hao Shi,\authormark{1,4,†} Yuhan Bao, \authormark{1,†}  Zhenshan Bing,\authormark{4} Yiyi Liao,\authormark{3} Kailun Yang,\authormark{2,*} and Kaiwei Wang\authormark{1,*}}

\address{\authormark{1}State Key Laboratory of Modern Optical Instrumentation, Zhejiang University, China\\
\authormark{2}School of Robotics and National Engineering Research Center of Robot Visual Perception and Control Technology, Hunan University, China\\
\authormark{3}College of Information Science and Electronic Engineering, Zhejiang University, China\\
\authormark{4}The Chair of Robotics, AI, and Real-Time Systems, Technical University of Munich, Germany\\
\authormark{†}These authors contributed equally.}

\email{\authormark{*}wangkaiwei@zju.edu.cn, kailun.yang@hnu.edu.cn} 

\begin{abstract*} 
Achieving 3D reconstruction from images captured under optimal conditions has been extensively studied in the vision and imaging fields. However, in real-world scenarios, challenges such as motion blur and insufficient illumination often limit the performance of standard frame-based cameras in delivering high-quality images. To address these limitations, we incorporate a transmittance adjustment device at the hardware level, enabling event cameras to capture both motion and exposure events for diverse 3D reconstruction scenarios. Motion events (triggered by camera or object movement) are collected in fast-motion scenarios when the device is inactive, while exposure events (generated through controlled camera exposure) are captured during slower motion to reconstruct grayscale images for high-quality training and optimization of event-based 3D Gaussian Splatting (3DGS). Our framework supports three modes: High-Quality Reconstruction using exposure events, Fast Reconstruction relying on motion events, and Balanced Hybrid optimizing with initial exposure events followed by high-speed motion events. On the EventNeRF dataset, we demonstrate that exposure events significantly improve fine detail reconstruction compared to motion events and outperform frame-based cameras under challenging conditions such as low illumination and overexposure. Furthermore, we introduce EME-3D, a real-world 3D dataset with exposure events, motion events, camera calibration parameters, and sparse point clouds. Our method achieves faster and higher-quality reconstruction than event-based NeRF and is more cost-effective than methods combining event and RGB data. By combining motion and exposure events, E-3DGS sets a new benchmark for event-based 3D reconstruction with robust performance in challenging conditions and lower hardware demands. The source code and dataset will be available at \url{https://github.com/MasterHow/E-3DGS}.

\end{abstract*}

%%%%%%%%%%%%%%%%%%%%%%%%%%  body  %%%%%%%%%%%%%%%%%%%%%%%%%%
\section{Introduction}
Reconstructing 3D scenes and objects from images has been a fundamental research focus in various vision applications, including imaging systems~\cite{kim2010principles, park2018quantitative}, depth perception~\cite{schwarz2010mapping}, virtual reality~\cite{xu2023vr, laviola2008bringing}, and high-precision measurements~\cite{zhang2018high, geng2011structured}.
Neural Radiance Field (NeRF)~\cite{mildenhall2021nerf} has become a powerful tool for 3D reconstruction, but its limited training and rendering efficiency hinder real-time applications.
3D Gaussian Splatting (3DGS)~\cite{kerbl20233d} advances NeRF by using explicit 3D Gaussians and tile-based rasterization, enhancing both efficiency and synthesis quality.

Despite the advancements in NeRF~\cite{mildenhall2021nerf} and 3DGS~\cite{kerbl20233d}, it is notable that both methods heavily rely on high-quality training images to achieve accurate 3D reconstructions with RGB cameras~\cite{klenk2023nerf}. 
Unfortunately, motion blur is a common challenge in real-world systems, particularly in fast-moving scenarios or low-light conditions where longer exposure times are necessary~\cite{klenk2023nerf, gallego2020event, rudnev2023eventnerf}. 
This blur often leads to mismatched feature points in COLMAP~\cite{snavely2006photo}, resulting in inaccuracies in pose calibration and point cloud initialization, and sometimes even causing failures in recovering camera poses, thereby hampering the training of 3DGS.

Traditional frame cameras capture images by integrating light over a fixed exposure time, but they suffer from limited dynamic range, often losing details in highlights or shadows~\cite{yasuma2010generalized, nayar2000high}.
In contrast, event cameras~\cite{gallego2020event}, which capture log-intensity changes asynchronously with high temporal resolution, provide a promising solution to these challenges.
For example, in the digital preservation of cultural heritage, event cameras can capture subtle surface details of artifacts under low-light conditions, minimizing the risk of damage from excessive illumination~\cite{gomes20143d}. 
Furthermore, their high temporal resolution facilitates the rapid acquisition of 3D models, crucial for digitizing valuable assets quickly and efficiently in dynamic environments~\cite{xu2023vr}. 
Their low power consumption also makes them well-suited for industrial applications like long-term environmental monitoring or remote robotics, minimizing environmental impact~\cite{klenk2023nerf, rudnev2023eventnerf}.
Several approaches~\cite{klenk2023nerf, rudnev2023eventnerf, hwang2023ev, qi2023e2nerf} have combined event cameras with NeRF to enhance 3D reconstruction in difficult environments. 
However, achieving real-time high-fidelity rendering remains a significant challenge for these implicit methods. 
Recent efforts~\cite{xiongevent3dgs, wang2024evggs, yu2024evagaussians} have explored integrating event cameras with 3D Gaussian Splatting (3DGS). 
However, these approaches overlook the source of event generation, often relying on additional RGB sensors for fusion or learning-based methods~\cite{Rebecq19cvpr} to convert events into pseudo-grayscale images to achieve high-quality reconstructions, making it challenging to obtain such results using a single event sensor.

To achieve high-quality 3D reconstruction using only a single event sensor, we combine motion and exposure events to balance quality and efficiency in diverse scenarios.
We introduce three operating modes for E-3DGS to address varying scene reconstruction needs:
\begin{itemize}
    \item \textbf{High-Quality Reconstruction Mode}: 
    In this mode, a programmable variable aperture gradually increases sensor brightness. 
    The time each pixel reaches target brightness is recorded, generating grayscale images to guide 3DGS for high-quality reconstruction.
    \item \textbf{Fast Reconstruction Mode}: 
    Utilizing a High-Definition (HD) resolution event camera to capture motion events for supervising 3DGS in high-speed scenes.
    \item \textbf{Balanced Hybrid Mode}: This mode starts with a few exposure events captured by the variable aperture, followed by motion events captured automatically. During 3DGS optimization, both exposure images and motion events jointly supervise the process, balancing reconstruction speed and quality in challenging scenarios.
\end{itemize}

On the EventNeRF dataset~\cite{rudnev2023eventnerf}, we simulate exposure events and map them to grayscale images, demonstrating their advantages ($35.73$ dB vs. $17.29$ dB) over motion events combined with E2VID in the 3DGS pipeline.
By incorporating a programmable variable aperture into event cameras, we enhance their ability to capture fine spatial details, albeit at the cost of reduced effectiveness in mitigating motion blur in high-speed scenarios. 
Additionally, this method of collecting exposure events retains a significantly higher dynamic range compared to conventional RGB cameras.
Under extreme lighting conditions, the advantages of exposure events become particularly evident. 
In low-light environments, RGB cameras fail to capture contours and other spatial details, whereas grayscale images derived from exposure events maintain high image quality.
Similarly, in cases of overexposure, RGB cameras suffer from pixel saturation, losing crucial details in high-brightness regions due to their limited dynamic range, whereas exposure event-based grayscale images effectively preserve and recover fine details in these areas.
These results demonstrate the potential of incorporating a programmable variable aperture at the hardware level, combined with a temporal mapping method, to overcome the limitations of conventional RGB cameras and motion-event-only approaches, enabling robust 3D reconstruction using a single event sensor in challenging optical scenarios.

Moreover, to verify the effectiveness of the proposed E-3DGS, we established a hardware acquisition system using a programmable variable aperture and a high-resolution event camera to create \textbf{EME-3D}, the first real-world 3D reconstruction dataset which distinguishes between \textbf{E}xposure and \textbf{M}otion \textbf{E}vents. As shown in Tab.~\ref{tab:Dataset_comparison}, EME-3D contains nine sequences of $1280{\times}720$ high-resolution event streams, camera calibration parameters, sparse point clouds, and exposure events for reconstructing high-quality grayscale images. Experimental results demonstrate that, compared to event-based NeRF and event-to-grayscale learning-based 3DGS methods, E-3DGS achieves faster reconstruction, higher quality, and greater flexibility in handling diverse scene requirements.
In Fast Reconstruction Mode, E-3DGS achieves a PSNR gain of $5.68dB$ over EventNeRF, along with a significantly higher rendering speed ($79.37$ FPS \textit{vs.} $0.03$ FPS). 
In the High-Quality Reconstruction Mode, E-3DGS delivers a PSNR increase of $10.89dB$ compared to the event-to-grayscale learning-based 3DGS baseline.

\begin{table}[!t]
    \centering
    \caption{Comparison of Event-based 3D Reconstruction Datasets.}
    \begin{adjustbox}{width=0.6\columnwidth}
    {
    \begin{tabular}{lccc}
    \toprule
    \textbf{Feature} & \textbf{EDS~\cite{hidalgo2022event}} & \textbf{EventNeRF~\cite{rudnev2023eventnerf}} & \textbf{EME-3D (Ours)} \\
    \midrule
    \textbf{Number of scenes} & 16 & 7 & 9 \\
    \textbf{Image resolution} & 346$\times$260 & 346$\times$260 & 1280$\times$720 \\
    \textbf{Modality} & RGB, Event & RGB, Event & Event \\
    \textbf{Motion events} & \CheckmarkBold & \CheckmarkBold & \CheckmarkBold \\
    \textbf{Exposure events} & \XSolidBrush & \XSolidBrush & \CheckmarkBold \\
    \textbf{Number of images} & 1,030 & 1,000 & 200 \\
    \textbf{Sharp images} & \XSolidBrush & \CheckmarkBold & \CheckmarkBold \\
    \textbf{Real data} & \CheckmarkBold & \XSolidBrush & \CheckmarkBold \\
    \bottomrule
    \end{tabular}
    }
    \end{adjustbox}
    \label{tab:Dataset_comparison}
\end{table}

The main contributions can be summarized as follows:

\begin{itemize}
    \item This work is the first to incorporate exposure event information into event-based 3D Gaussian Splatting (3DGS), converting sparse events during exposure into dense intensity frames for high-quality event-based 3D reconstruction.
    \item To the best of our knowledge, EME-3D is the first real-world 3D dataset that contains both exposure and motion events, camera parameters, and sparse 3D point clouds, and we make it publicly available.
    \item A hybrid approach is proposed, capturing exposure events at low speeds and motion events at high speeds, effectively balancing reconstruction quality and efficiency.
\end{itemize}

\section{Related work}
\subsection{Neural 3D Reconstruction}
Traditional 3D reconstruction methods, such as point clouds~\cite{achlioptas2018learning}, meshes~\cite{liu2020general}, and voxel grids~\cite{lombardi2019neural, sitzmann2019deepvoxels}, rely on explicit representations but are limited by their fixed topological structures. NeRFs~\cite{mildenhall2021nerf} have gained traction for synthesizing high-quality novel views using MLP-based neural networks and differentiable volume rendering, but they suffer from low rendering efficiency and long training times. 
More recently, 3D Gaussian Splatting (3DGS)~\cite{kerbl20233d} has demonstrated a significant advancement, offering faster convergence and superior rendering quality by employing Gaussian splats.
However, all of these methods require clear RGB inputs and struggle with the fast motion or low-light conditions commonly encountered in real-world applications.

\subsection{Event-based 3D Reconstruction}
Event cameras~\cite{gallego2020event}, which detect asynchronous brightness changes at the pixel level, offer substantial benefits in high dynamic range and in scenarios involving high-speed motion or rapid dynamic changes.
Recent works have leveraged the complementary strengths of event streams and traditional RGB frames to enhance 3D scene reconstruction~\cite{klenk2023nerf, qi2023e2nerf, yu2024evagaussians}. 
Incorporating event data into NeRF-based models~\cite{mildenhall2021nerf}, such as Ev-NeRF~\cite{hwang2023ev} and EventNeRF~\cite{rudnev2023eventnerf}, has demonstrated the potential for multi-view consistency in 3D reconstruction. 
However, NeRFs' computational demands hinder real-time rendering. To address this, event-based 3D Gaussian Splatting (3DGS)~\cite{xiongevent3dgs,wu2024ev} offers a more efficient alternative, accelerating reconstruction while preserving high-quality geometry and appearance. Nevertheless, due to limited texture in event streams, previous event-based 3DGS reconstructions often lack fine-grained detail~\cite{xiongevent3dgs,wu2024ev}. Therefore,
some studies have attempted to overcome this by using learning-based event-to-image methods~\cite{wang2024evggs,Rebecq19cvpr}, or event-image deblurring methods~\cite{deguchi2024e2gs, weng2024eadeblur} to enhance the reconstruction accuracy. However, learning-based approaches require additional computational resources and yield limited quality improvements, while event-image modal fusion increases hardware complexity, size, and cost.
In earlier work, EvTemMap~\cite{bao2024temporal} distinguished an event-based photography method by converting the precise timing of exposure events into dense grayscale frames, thereby eliminating the need for multiple sensors. 
Instead of introducing learning-based event-to-image or event-image deblurring modules, we propose to explore exposure and motion events in 3D explicit reconstruction, which enhances both the quality and speed of reconstruction from pure event data, particularly beneficial in low-light or high-dynamic-range scenarios.

\begin{figure}[!t]
\centering
\includegraphics[width=1.0\linewidth]{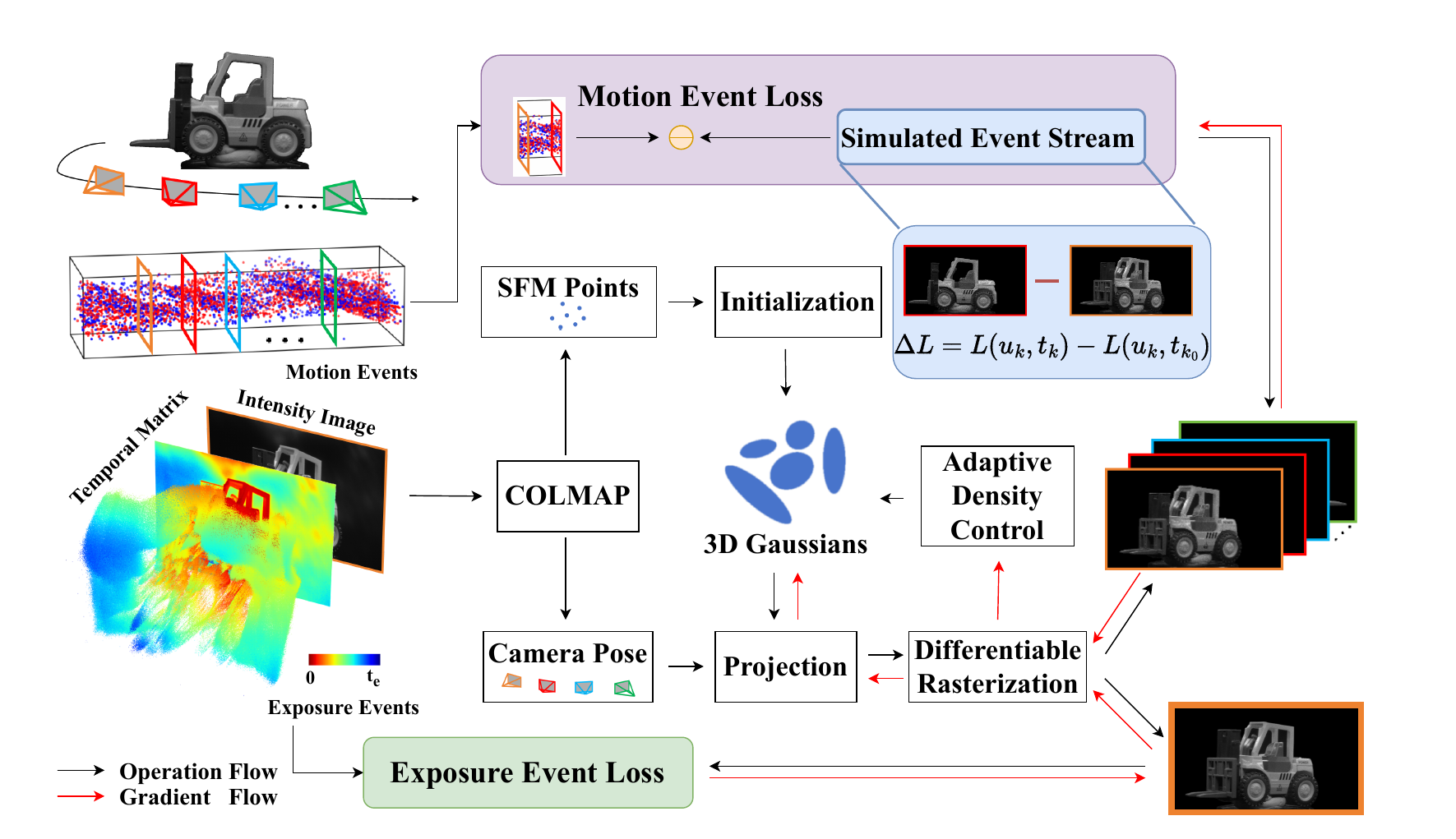}
\caption{
Overview of the proposed E-3DGS framework. This framework integrates motion and exposure events for training 3DGS to effectively handle diverse real-world conditions. We utilize Temporal-to-Intensity Mapping to convert exposure events into intensity images, which yield camera trajectories and a sparse point cloud for 3DGS training. The optimization of 3DGS parameters is supervised through motion event loss and exposure event loss.}
\label{Flow}
\end{figure}

\section{Methodology}
In this section, we present E-3DGS, an event-based method that partitions events into motion and exposure for training and optimization of 3D Gaussian Splatting (3DGS), as illustrated in Fig.~\ref{Flow}. 
Exposure events are mapped into grayscale images, which are then processed using COLMAP~\cite{snavely2006photo} to perform Structure from Motion (SfM), a method that estimates camera poses and initializes sparse 3D point clouds based on 2D image sequences.
This step is necessary for real-world datasets, while in the EventNeRF dataset~\cite{rudnev2023eventnerf}, camera poses are provided, and the initialization is performed randomly. 
Subsequently, the real-world motion events and grayscale images derived from exposure events are used as supervision signals, enabling 3D reconstruction tailored to different scene requirements through three operational modes. 
Our approach utilizes the high temporal resolution of motion events and the rich texture information from exposure events, effectively integrating both capabilities within a single event camera.
In the preliminary section (Sec.~\ref{Method:pre}), we introduce the 3DGS framework and the event camera model. 
Next, we detail the process of mapping temporal information from exposure events into high-quality grayscale images (Sec.~\ref{Method:temporal_mapping}), the method for selecting the event window (Sec.~\ref{Method: event_window_selection}), and the formulation of the overall loss function (Sec.~\ref{Method:loss_functions}).

\subsection{Preliminary: 3D Gaussian Splatting \& Event Model}
\label{Method:pre}

In this section, we introduce the foundational elements of our approach, which combines 3D Gaussian Splatting (3DGS) and the event camera model.

\textbf{3D Gaussian Splatting (3DGS)} represents a 3D scene using anisotropic 3D Gaussians, each defined by a mean \(\mu \in \mathbb{R}^3\), a covariance matrix \(\Sigma \in \mathbb{R}^{3 \times 3}\), and an opacity \(\alpha\). The covariance matrix \(\Sigma\) is factorized as:
\begin{equation}
\Sigma = RS S^\top R^\top,
\label{eq:covariance_decomposition}
\end{equation}
where \(R\) is a rotation matrix and \(S\) is a scaling matrix, ensuring the matrix remains positive semi-definite for optimization. For rendering, the 3D Gaussians are projected onto the 2D image plane, with the covariance matrix transformed into camera coordinates as:
\begin{equation}
\Sigma' = J W \Sigma W^\top J^\top,
\label{eq:projected_covariance}
\end{equation}
where \(W\) is the view transformation, and \(J\) is the Jacobian of the projective transformation. Colors are computed via \(\alpha\)-blending to combine \(N\) Gaussians as:
\begin{equation}
C(u) = \sum_{i=1}^{N} c_i \alpha_i \prod_{j=1}^{i-1} (1 - \alpha_j),
\label{eq:alpha_blending_color}
\end{equation}
where \(c_i\) is the color defined by spherical harmonics, and \(\alpha_i\) is the multiplication of opacity and the transformed 2D Gaussian. 
Despite its effectiveness in scene reconstruction and novel view synthesis, 3DGS struggles with real-world conditions such as motion blur or low-light conditions.

\textbf{Event Camera Model}: Each event \(e_k = (u_k, t_k, p_k)\) captures the pixel coordinate \(u_k\), timestamp \(t_k\), and polarity \(p_k\), reflecting a brightness change exceeding the contrast threshold \(C\). The brightness change for each event is given by:
\begin{equation}
\Delta L = L(u_k, t_k) - L(u_k, t_{k-1}) = p_k \cdot C,
\label{eq:delta_L}
\end{equation}
where \(t_{k-1}\) represents the timestamp of the previous event at the same pixel. The instantaneous intensity image at time \(t\), denoted as \(I(t) \in \mathbb{R}^{W \times H}\), undergoes a logarithmic transformation to better represent perceptual brightness changes:
\begin{equation}
L(t) = \frac{\log I(t)}{g},
\end{equation}
where \(g = 2.2\) is a gamma correction factor applied in all experiments to linearize intensity values. This correction compensates for non-linear intensity encoding intended for display on sRGB monitors, with \(g = 2.2\) providing an effective approximation for standard gamma correction~\cite{iec_standard}.

\subsection{Temporal-to-Intensity Mapping of Exposure Events}
\label{Method:temporal_mapping}
Due to the limited texture information provided by motion events, we propose capturing exposure events during scene reconstruction by controlling the camera's aperture. High-quality grayscale images are then generated through temporal-to-intensity mapping of these exposure events, achieved by dynamically adjusting the event camera's transmittance.
Specifically, we introduce a Transmittance Adjustment (TA) device, where the transmittance rate \( TR(t) \) changes from \(0\) to \(1\) according to the function:
\begin{equation}
TR(t) = 
\begin{cases} 
f(t), & 0 \leq t \leq t_{\text{end}} \\
1, & t > t_{\text{end}} 
\end{cases}
\label{eq:transmittance}
\end{equation}
where \(t_{\text{end}}\) represents the end time of the adjustment period, and \(f(t)\) is a predefined function governing the transition of transmittance over time.
Each pixel triggers an event at a specific time \( t^*(u) \), forming a temporal matrix. 
The intensity-proportional value $I_{max}(u)$, which represents the scene's intensity, can be derived from the event time $t^*(u)$ as follows:
\begin{equation}
I_{max}(u) = \frac{C}{h(t^*(u))},
\label{eq:Imax_corrected} %
\end{equation}
where 
\begin{equation}
h(t^*) = \int_0^{t^*(u)} TR(t) \, dt,
\label{eq:integral}
\end{equation}
where \( t^*(u) \) represents the timestamp at which the initial positive event (IPE) is triggered for the pixel at \(u = (x, y)\). 
Here, $C$ is a constant representing the threshold charge required to trigger an event, incorporating sensor parameters like the photodiode capacitance ($C_{PD}$), reference voltage ($V_{ref}$), and threshold voltage ($V_{thd}$).
After mapping the time information $t^*(u)$ to intensity-proportional grayscale values $I_{max}(u)$ using Eq. \eqref{eq:Imax_corrected}, normalization is performed by adjusting $I_{max}(u)$ to the range $[0, 1]$, yielding a high-resolution grayscale image with adaptive dynamic range. This normalization step effectively handles the pixel-specific constant $C$.
The process is robust to various forms of degradation (including noise, contrast threshold variability, and pixel anomalies) because its degradation model explicitly simulates and incorporates these specific real-world imperfections during training data generation, thereby enabling the final restoration model to learn how to handle and mitigate them, as detailed in prior work~\cite{bao2024temporal}.
The resulting temporal-to-intensity mapping of exposure events can further introduce additional constraints to optimize the event-based 3DGS, as detailed in Sec.~\ref{Method:loss_functions}.

\subsection{Event Window Selection}
\label{Method: event_window_selection}
As shown in Fig.~\ref{Flow}, we select two viewpoints to compute the motion event loss, leveraging the events occurring between them as supervision signals. 
To achieve this, we first define the start and end points of the event windows. Specifically, the event stream is divided into intervals represented by $t_i = i / N_\text{windows}$, where $i \in {1, \ldots, N_\text{windows}}$ and $t_i$ denotes the window endpoints.
Fixed-length windows pose significant challenges, as short intervals can impede the propagation of global illumination effects, while long intervals may degrade the fidelity of local details. 
To address this, we employ a randomized window sampling strategy~\cite{rudnev2023eventnerf}. 
Given a fixed window end $t$, the window start $t_0$ is sampled as $t_0 \sim \mathcal{U}[t - L_\text{max}, t)$, or equivalently, the window length is drawn as $t - t_0 \sim \mathcal{U}(0, L_\text{max}]$. 
This approach enables the model to learn both global lighting and fine-grained details, striking a balance between global coherence and local accuracy during reconstruction.

\subsection{Loss Functions}
\label{Method:loss_functions}
The loss function guiding the training process consists of two primary components: the motion event loss and the exposure event loss. 
The motion event loss is crucial for ensuring that the predicted brightness variations align with the motion events captured by the event camera. 
Based on the event window selection process detailed in Sec.~\ref{Method: event_window_selection}, the motion event loss is formulated using the events occurring within the time interval defined by the start timestamp $t_{k_0}$ and the end timestamp $t_k$ as:
\begin{equation}
L_{\text{evs}, \text{norm}} = \left\|\frac{\Delta \hat{L}(u_k)}{\|\Delta \hat{L}(u_k)\|_2} - \frac{\Delta L(u_{k})}{\|\Delta L(u_{k})\|_2}\right\|_2^2,
\label{eq:motion_event_loss}
\end{equation}
where \(\Delta \hat{L}(u_k)\) represents the predicted logarithmic brightness change, calculated as \( \Delta \hat{L}_k = \hat{L}(u_k,t_k) - \hat{L}(u_k, t_{k_0}) \). 
The predicted brightness values, \( \hat{I}(u_k, t_k) \) and \( \hat{I}(u_k, t_{k_0}) \), are obtained from the 3DGS model through volumetric rendering. 
This formulation ensures that the predicted brightness changes are consistent with the real-world motion event stream observed by the event camera, allowing for accurate motion capture under high-motion scenes.

In addition, the exposure event loss supervises the 3DGS reconstruction of high-quality frames obtained from a temporal-to-intensity mapping of exposure events.
It is computed as the $L_2$ loss (squared error) between the predicted image and the ground truth:
\begin{equation}
L_{\text{img}} = \frac{1}{N} \sum_{k=1}^{N} \left( I_{\text{pred}}(u_{\text{k}}) - I_{\text{gt}}(u_{\text{k}}) \right)^2,
\label{eq:exposure_event_loss_l2}
\end{equation}
where \(I_{\text{pred}}(u_{\text{k}})\) and \(I_{\text{gt}}(u_{\text{k}})\) are the predicted and ground truth color values at pixel \(u_{\text{k}}\), respectively, and \(N\) represents the total number of pixels in the image. The final combined loss used during training is:
\begin{equation}
L = \lambda \cdot L_{\text{evs}, \text{norm}} + (1 -\lambda) \cdot L_{\text{img}},
\label{eq:total_loss}
\end{equation}
where \( \lambda = 0 \) prioritizes high-quality texture reconstruction ( High-Quality Reconstruction Mode), \( \lambda = 1 \) focuses on fast motion capture (Fast Reconstruction Mode), and \( \lambda = 0.5 \) balances both for a compromise between speed and quality (Balanced Hybrid Mode).
By incorporating exposure events into the 3DGS framework, this loss adds a novel constraint that enhances texture reconstruction from sparse event data using a single event sensor. 

\section{Experiments}

\subsection{Evaluation Dataset}
\label{Experiments:Evaluation Dataset}

\textbf{Synthetic EventNeRF Dataset.} 
The EventNeRF~\cite{rudnev2023eventnerf} dataset provides RGB images and corresponding motion event streams for seven sequences, featuring a constant gray background. 
To maintain a consistent background intensity across each sequence, we first segment the RGB images using EfficientSAM~\cite{xiong2023efficientsam}, as our focus is on 3D reconstruction of foreground objects.
To realistically simulate varying lighting, we scale the segmented foreground intensities by factors of $0.25$ for low-light and $1.5$ for overexposure.
The segmented foreground is then normalized into $100$ intensity levels ranging from $0$ to $1$, simulating the aperture’s transition from fully closed to fully open. 
These intensity-modulated images are processed through DVS-Voltmeter~\cite{lin2022dvsvoltmeter} to generate exposure event streams. 
The resulting event streams are converted into grayscale images using the Temporal-to-Intensity Mapping method described in Section~\ref{Method:temporal_mapping}. 
Finally, the background is restored to its designated intensity value using the segmentation masks, ensuring photometric consistency across the dataset.

\begin{figure}[!t]
\centering
\includegraphics[width=0.5\linewidth]{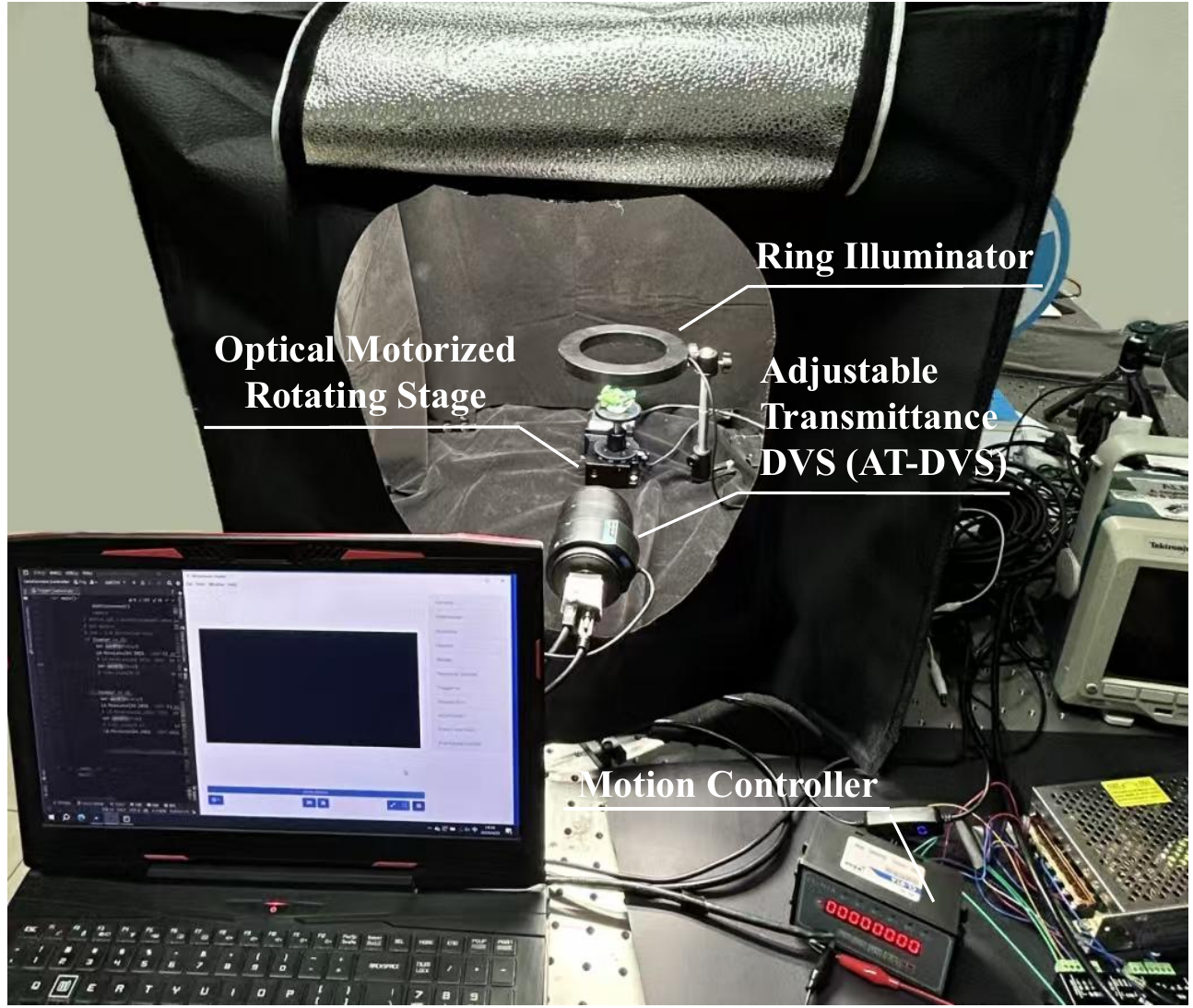}
\caption{Real-world data acquisition setup: The object is placed on a motorized optical rotation stage and illuminated by an overhead ring light to ensure uniform lighting. The scene is captured using an AT-DVS, implemented with a Prophesee Evaluation Kit 4 HD (EVK4) and an aperture shutter, facilitating high dynamic range event-based grayscale imaging through temporal-to-intensity mapping by exposure events.}
\label{fig:Acquisition}
\end{figure}

\textbf{EME-3D Real-World Dataset.}
We utilize an AT-DVS (Adjustable Transmittance DVS), specifically the Metavision Evaluation Kit 4 HD by Prophesee equipped with an aperture shutter placed in front of the lens, to capture real-world sequences. 
Fig.~\ref{fig:Acquisition} illustrates the experimental setup used for data acquisition. 
The setup incorporates an optical motorized rotating stage, specifically the YXRA100 rotational table (Beijing Yixuan Yunkong Technology Co., Ltd.; product details in \cite{YXRA60}), offering a rotational precision of $0.005^\circ$.
The imaging system features a $9{\sim}50mm$ LensConnect BH Series variable focal length lens (Edmund Optics, product number $53086$, model EL6Z0915UCS-MPWIR), providing a horizontal field of view (FoV) of $9.18^\circ$ and a vertical FoV of $7.78^\circ$ during data acquisition.
Before the experiment, the object is securely affixed to the stage using Blu-Tack to prevent any relative movement during rotation. 
Each object undergoes two recording phases, with the rotating stage reset to the zero position before each phase. 
In the first phase, the stage performs a full $360${\textdegree} rotation to capture motion events. 
During the second phase, the stage rotates in $1.8${\textdegree} increments, controlled by a motion controller, to capture exposure events at specific positions.

\subsection{Implementation Details}
Our implementation is based on the official 3D-GS codebase~\cite{kerbl20233d}. 
For training in the fast reconstruction mode, which relies solely on motion event supervision, we limit the training schedule to $10,000$ iterations to mitigate the risk of overfitting caused by the sparse spatial detail of motion events. 
This training includes a $500$-iteration warmup phase with an initial learning rate of $1.6 \times 10^{-5}$.
For the high-quality reconstruction mode (which utilizes only exposure events for supervision) and the balanced hybrid mode (where supervision transitions from exposure events during slow motion to motion events during fast motion), we run for $30,000$ iterations with a $500$-iteration warm-up and an initial learning rate of $1.6 \times 10^{-4}$. 
In the EventNeRF dataset~\cite{rudnev2023eventnerf}, the ground truth for quantitative evaluation is derived by directly converting the RGB images provided in the dataset to grayscale. 
For our real-world dataset, the grayscale images mapped from exposure events are used as the ground truth for evaluation.
All experiments were performed on a single NVIDIA RTX 3090-Ti GPU.

\subsection{Baselines}
\label{Exp:baseline}
We benchmark our approach against a NeRF-based method, EventNeRF~\cite{rudnev2023eventnerf}, and a naive baseline, E2VID~\cite{Rebecq19cvpr} + NeRF~\cite{mildenhall2021nerf}, which cascades the event-to-video pipeline E2VID into NeRF. Additionally, we compare a method that cascades E2VID with a vanilla 3D Gaussian Splatting approach~\cite{kerbl20233d}. 
We observed significant background noise in the images reconstructed by the learning-based E2VID, which adversely affects the accuracy of E2VID + 3DGS. 
To address this, we have introduced a stronger baseline by introducing the Segment Anything Model (SAM)~\cite{xiong2023efficientsam} to segment and remove background noise from the grayscale images reconstructed by E2VID. The segmentation quality was manually double-checked and refined, forming our strongest baseline: E2VID + SAM + 3DGS.
For the EventNeRF~\cite{rudnev2023eventnerf} dataset, we convert the color results provided by EventNeRF~\cite{rudnev2023eventnerf} to grayscale directly for comparison.
For our collected real-world scenes, we retrained EventNeRF and all baselines on the EME-3D dataset, rendering RGB and depth images for comparison to ensure fairness and reliability.

\subsection{Quantitative Comparison}
To evaluate the feasibility of utilizing exposure events for 3D reconstruction, as detailed in Section~\ref{Experiments:Evaluation Dataset}, we first simulate exposure events from RGB images and subsequently generate grayscale images using the Temporal-to-Intensity Mapping method. 
As shown in Tab.~\ref{tab:comparison_synthetic}, leveraging exposure events under the 3DGS framework increases PSNR from $17.29$ (E2VID + 3DGS) to $35.73$. 
This demonstrates the advantages of hardware-augmented event cameras in generating intensity images over learning-based methods.
Across the first six sequences, 3D reconstruction driven by exposure events consistently surpasses the other three approaches in PSNR, SSIM, and LPIPS, demonstrating its robustness and effectiveness. 
In the mic sequence, EventNeRF~\cite{rudnev2023eventnerf} slightly outperforms our method due to segmentation errors during exposure event simulation. 
EfficientSAM~\cite{xiong2023efficientsam} faced challenges as the mic sequence's small foreground and its similarity to the background increased segmentation errors, degrading grayscale reconstruction. 
Nonetheless, our results highlight the potential of exposure events to enhance event cameras by leveraging temporal intensity dynamics for robust 3D reconstruction.

As shown in Tab.~\ref{tab:comparison}, we further compare the proposed E-3DGS method with other baselines (see Sec.~\ref{Exp:baseline}) on the real-world EME-3D dataset. In fast reconstruction mode, E-3DGS outperforms all baselines across all sequences in terms of reconstruction quality. On average, E-3DGS (Fast Reconstruction Mode), using only motion events, achieves a PSNR improvement of \textbf{$2.19dB$} ($25.22$ \textit{vs.} $23.03$) over the strongest baseline E2VID + SAM + 3DGS, which utilizes additional learning-based events-to-video~\cite{Rebecq19cvpr} and vision foundation model~\cite{xiong2023efficientsam} techniques. 
Moreover, compared to the classic event-only method EventNeRF~\cite{rudnev2023eventnerf}, E-3DGS (Fast Reconstruction Mode) shows a PSNR improvement of \textbf{$5.68dB$} ($25.22$ \textit{vs.} $19.54$). 
In high-quality reconstruction mode, E-3DGS fully leverages the high-resolution spatial cues from exposure events to optimize the Gaussian ellipsoids and distributions, resulting in a substantial further improvement in reconstruction quality. 
On average, E-3DGS achieves a PSNR increase of \textbf{$10.89dB$} ($33.92$ \textit{vs.} $23.03$), an SSIM improvement of \textbf{$0.08$} ($0.97$ \textit{vs.} $0.89$), and an LPIPS reduction of \textbf{$0.03$} ($0.09$ \textit{vs.} $0.12$) compared to the strongest E2VID + SAM + 3DGS baseline.
Notably, the training process of E-3DGS does not rely on the complex computations required by the learning-based E2VID or SAM.
Additionally, E-3DGS demonstrates a significant advantage in rendering speed achieving $79.37$ FPS, whereas EventNeRF renders at only $0.03$ FPS (our evaluation) or $0.045$ FPS (as reported in~\cite{wang2024evggs} as a baseline comparison).

\begin{table*}[!t]
    \centering
    \caption{Quantitative Comparisons on the Synthetic Dataset.}
    \begin{adjustbox}{width=\textwidth}
    {\small{
    \begin{tabular}{llcccccccc}
    \toprule
    & & \multicolumn{8}{c}{\textbf{Synthetic EventNeRF Dataset}} \\
    \cmidrule(lr){3-10}
    \raisebox{1.5ex}[0pt]{\textbf{Metric}} & \raisebox{1.5ex}[0pt]{\textbf{Method}} & chair & drums & ficus & hotdog & lego & materials & mic & Average \\
    \midrule
    \multirow{4}{*}{\textbf{PSNR (dB)}} 
    & E2VID~\cite{Rebecq19cvpr, Rebecq19pami}+NeRF~\cite{mildenhall2021nerf} & 18.00 & 14.01 & 17.59 & 23.25 & 15.75 & 16.79 & 15.91 & 17.33 \\
    & EventNeRF~\cite{rudnev2023eventnerf} & 29.06 & 26.91 & 31.80 & 29.97 & 23.17 & 21.30 & \textbf{31.40} & 27.66 \\
    & E2VID~\cite{Rebecq19cvpr, Rebecq19pami}+3DGS~\cite{kerbl20233d} & 17.85 & 14.47 & 17.27 & 23.17 & 15.68 & 16.77 & 15.82 & 17.29 \\
    & \textbf{Ours (Exposure Events)} & \textbf{40.87} & \textbf{34.02} & \textbf{40.00} & \textbf{40.08} & \textbf{27.83} & \textbf{35.96} & 31.36 & \textbf{35.73}  \\
    \midrule
    \multirow{4}{*}{\textbf{SSIM $\uparrow$}} 
    & E2VID~\cite{Rebecq19cvpr, Rebecq19pami}+NeRF~\cite{mildenhall2021nerf} & 0.92 & 0.83 & 0.91 & 0.94 & 0.81 & 0.90 & 0.92 & 0.89 \\
    & EventNeRF~\cite{rudnev2023eventnerf} & 0.94 & 0.92 & 0.94 & 0.95 & 0.90 & 0.94 & \textbf{0.97} & 0.94 \\
    & E2VID~\cite{Rebecq19cvpr, Rebecq19pami}+3DGS~\cite{kerbl20233d} & 0.92 & 0.87 & 0.91 & 0.94 & 0.84 & 0.91 & 0.93 & 0.90 \\
    & \textbf{Ours (Exposure Events)} & \textbf{0.99} & \textbf{0.98} & \textbf{0.99} & \textbf{0.99} & \textbf{0.95} & \textbf{0.99} & \textbf{0.97} & \textbf{0.98}  \\
    \midrule
    \multirow{4}{*}{\textbf{LPIPS $\downarrow$}} 
    & E2VID~\cite{Rebecq19cvpr, Rebecq19pami}+NeRF~\cite{mildenhall2021nerf} & 0.13 & 0.24 & 0.16 & 0.11 & 0.28 & 0.16 & 0.15 & 0.18 \\
    & EventNeRF~\cite{rudnev2023eventnerf} & 0.06 & 0.07 & 0.05 & 0.05 & 0.13 & 0.07 & \textbf{0.03} & 0.07 \\
    & E2VID~\cite{Rebecq19cvpr, Rebecq19pami}+3DGS~\cite{kerbl20233d} & 0.12 & 0.17 & 0.16 & 0.11 & 0.19 & 0.16 & 0.12 & 0.15 \\
    & \textbf{Ours (Exposure Events)} & \textbf{0.01} & \textbf{0.02} & \textbf{0.01} & \textbf{0.01} & \textbf{0.05} & \textbf{0.01} & 0.04 & \textbf{0.02}  \\
    \bottomrule
    \end{tabular}
    }}
    \end{adjustbox}
    \label{tab:comparison_synthetic}
\end{table*}

\begin{table*}[!t]
    \centering
    \caption{Quantitative Comparisons of Neural Reconstruction in Real World.}
    \begin{adjustbox}{width=\textwidth}
    {\small{
    \begin{tabular}{llccccccccccc}
    \toprule
    & & \multicolumn{10}{c}{\textbf{Real-World EME-3D Dataset}} \\
    \cmidrule(lr){3-12}
    \raisebox{1.5ex}[0pt]{\textbf{Metric}} & \raisebox{1.5ex}[0pt]{\textbf{Method}} & car & motorcycle & camera & cat & crab & forklift & racing car & tortoise & train & Average \\
    \midrule
    \multirow{6}{*}{\textbf{PSNR (dB)}} 
    & E2VID~\cite{Rebecq19cvpr, Rebecq19pami}+NeRF~\cite{mildenhall2021nerf} & 22.57 & 24.34 & 20.94 & 24.72 & 22.73 & 21.94 & 24.38 & 20.27 & 21.99 & 22.65 \\
    & EventNeRF~\cite{rudnev2023eventnerf} & 18.58 & 23.50 & 22.66 & 17.97 & 19.48 & 19.01 & 19.82 & 15.82 & 18.98 & 19.54  \\
    & E2VID~\cite{Rebecq19cvpr, Rebecq19pami}+3DGS~\cite{kerbl20233d} & 14.48 & 17.79 & 18.01 & 14.58 & 14.82 & 14.51 & 17.13 & 16.12 & 15.80 & 15.92  \\
    & E2VID~\cite{Rebecq19cvpr, Rebecq19pami}+SAM~\cite{xiong2023efficientsam}+3DGS~\cite{kerbl20233d} & 22.89 & 25.22 & 21.20 & 25.14 & 22.97 & 21.95 & 24.97 & 20.89 & 22.08 & 23.03  \\
    & Ours (Fast Reconstruction Mode) & 23.62 & 26.99 & 25.69 & 26.81 & 25.26 & 23.08 & 25.91 & 24.57 & 25.04 & 25.22  \\
    & \textbf{Ours (High-Qualiy Reconstruction Mode)} & \textbf{33.16} & \textbf{35.89} & \textbf{35.11} & \textbf{31.92} & \textbf{33.58} & \textbf{32.60} & \textbf{34.82} & \textbf{35.54} & \textbf{32.66} & \textbf{33.92}  \\
    \midrule
    \multirow{6}{*}{\textbf{SSIM $\uparrow$}} 
    & E2VID~\cite{Rebecq19cvpr, Rebecq19pami}+NeRF~\cite{mildenhall2021nerf} & 0.89 & 0.90 & 0.84 & 0.91 & 0.89 & 0.78 & 0.88 & 0.87 & 0.83 & 0.87 \\
    & EventNeRF~\cite{rudnev2023eventnerf} & 0.56 & 0.48 & 0.62 & 0.66 & 0.66 & 0.51 & 0.53 & 0.21 & 0.45 & 0.52  \\
    & E2VID~\cite{Rebecq19cvpr, Rebecq19pami}+3DGS~\cite{kerbl20233d} & 0.12 & 0.08 & 0.13 & 0.06 & 0.10 & 0.17 & 0.09 & 0.07 & 0.13 & 0.11  \\
    & E2VID~\cite{Rebecq19cvpr, Rebecq19pami}+SAM~\cite{xiong2023efficientsam}+3DGS~\cite{kerbl20233d} & 0.92 & 0.92 & 0.89 & 0.92 & 0.91 & 0.80 & 0.91 & 0.91 & 0.85 & 0.89  \\
    & Ours (Fast Reconstruction Mode) & 0.90 & 0.91 & 0.88 & 0.92 & 0.90 & 0.81 & 0.91 & 0.90 & 0.85 & 0.89  \\
    & \textbf{Ours (High-Qualiy Reconstruction Mode)} & \textbf{0.97} & \textbf{0.98} & \textbf{0.97} & \textbf{0.97} & \textbf{0.96} & \textbf{0.93} & \textbf{0.97} & \textbf{0.98} & \textbf{0.96} & \textbf{0.97}  \\
    \midrule
    \multirow{6}{*}{\textbf{LPIPS $\downarrow$}} 
    & E2VID~\cite{Rebecq19cvpr, Rebecq19pami}+NeRF~\cite{mildenhall2021nerf} & 0.12 & 0.10 & 0.17 & 0.10 & 0.11 & 0.21 & 0.13 & 0.12 & 0.15 & 0.13 \\
    & EventNeRF~\cite{rudnev2023eventnerf} & 0.40 & 0.30 & 0.34 & 0.34 & 0.33 & 0.45 & 0.53 & 0.83 & 0.49 & 0.45  \\
    & E2VID~\cite{Rebecq19cvpr, Rebecq19pami}+3DGS~\cite{kerbl20233d} & 0.63 & 0.63 & 0.60 & 0.64 & 0.62 & 0.60 & 0.61 & 0.62 & 0.57 & 0.61  \\
    & E2VID~\cite{Rebecq19cvpr, Rebecq19pami}+SAM~\cite{xiong2023efficientsam}+3DGS~\cite{kerbl20233d} & 0.12 & 0.09 & 0.14 & 0.10 & 0.11 & 0.22 & 0.10 & 0.09 & 0.15 & 0.12  \\
    & Ours (Fast Reconstruction Mode) & 0.14 & 0.10 & 0.15 & 0.10 & 0.12 & 0.24 & 0.11 & 0.11 & 0.20 & 0.14  \\
    & \textbf{Ours (High-Qualiy Reconstruction Mode)} & \textbf{0.10} & \textbf{0.06} & \textbf{0.07} & \textbf{0.08} & \textbf{0.09} & \textbf{0.15} & \textbf{0.06} & \textbf{0.06} & \textbf{0.10} & \textbf{0.09}  \\
    \bottomrule
    \end{tabular}
    }}
    \end{adjustbox}
    \label{tab:comparison}
\end{table*}

\begin{figure*}[!t]
\centering
\includegraphics[width=1.0\linewidth]{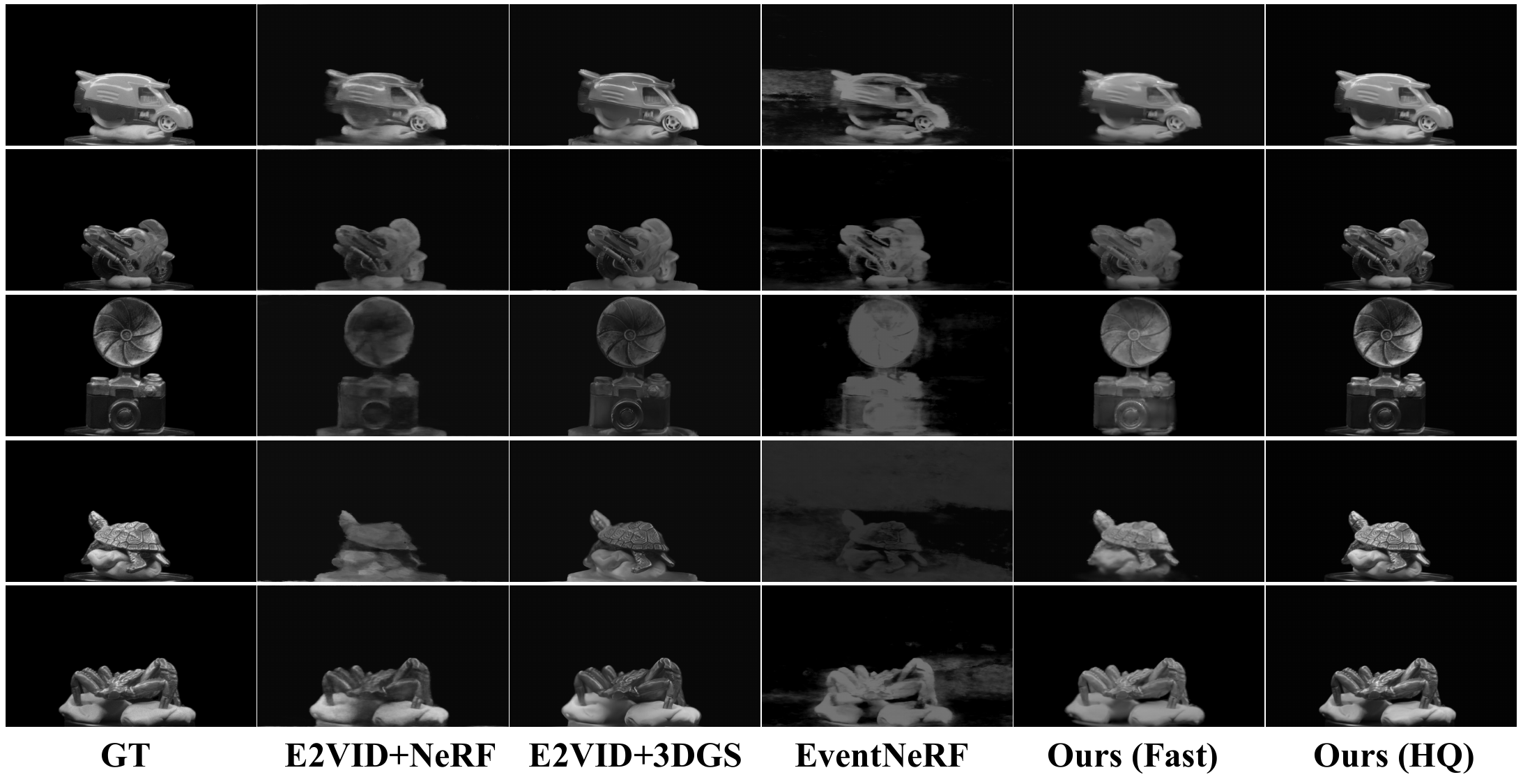}
\caption{Qualitative comparison of appearance reconstruction on the real-world EME-3D dataset. The E2VID + NeRF method successfully reconstructs the overall scene but lacks fine local details, such as the tortoise's shell texture. EventNeRF exhibits noticeable artifacts in both the car and the tortoise. In contrast, our proposed E-3DGS, in both fast and high-quality reconstruction modes, preserves sharper and more consistent structures while maintaining cleaner backgrounds. 
}
\label{fig:render_compare}
\end{figure*}

\begin{figure*}[!t]
\centering
\includegraphics[width=1.0\linewidth]{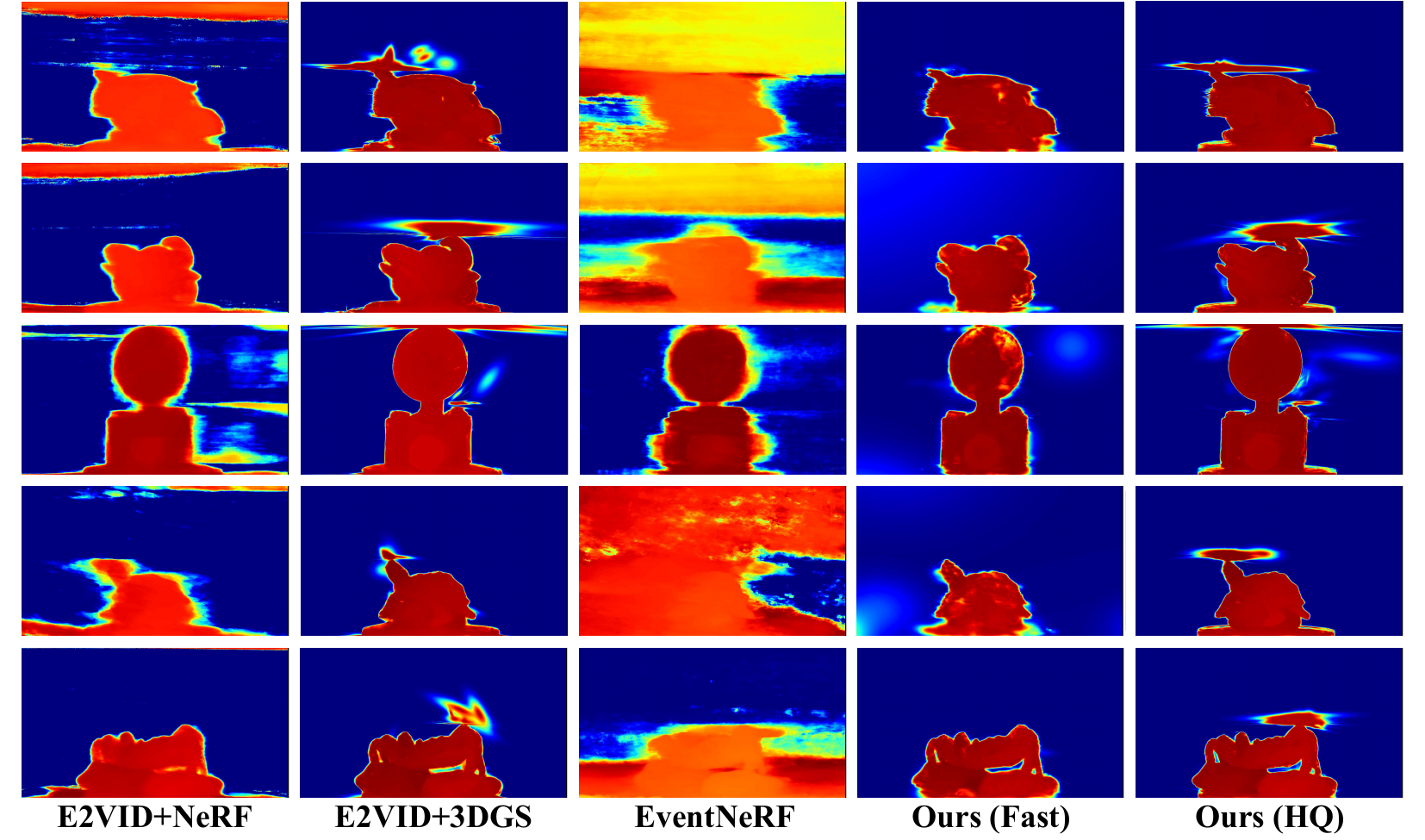}
\caption{Qualitative comparison of geometry reconstruction on the real-world EME-3D dataset. Both E2VID+NeRF and EventNeRF, which are based on NeRF, struggle to separate the foreground from the background and are affected by noticeable noise. In contrast, the physics-based 3DGS method handles geometry reconstruction more effectively.  Compared to E2VID+3DGS, our E-3DGS excels in capturing high-frequency spatial details, such as the gap between the crab’s body and the base.
}
\label{fig:depth_compare}
\end{figure*}

\begin{figure*}[!t]
\centering
\includegraphics[width=1.0\linewidth]{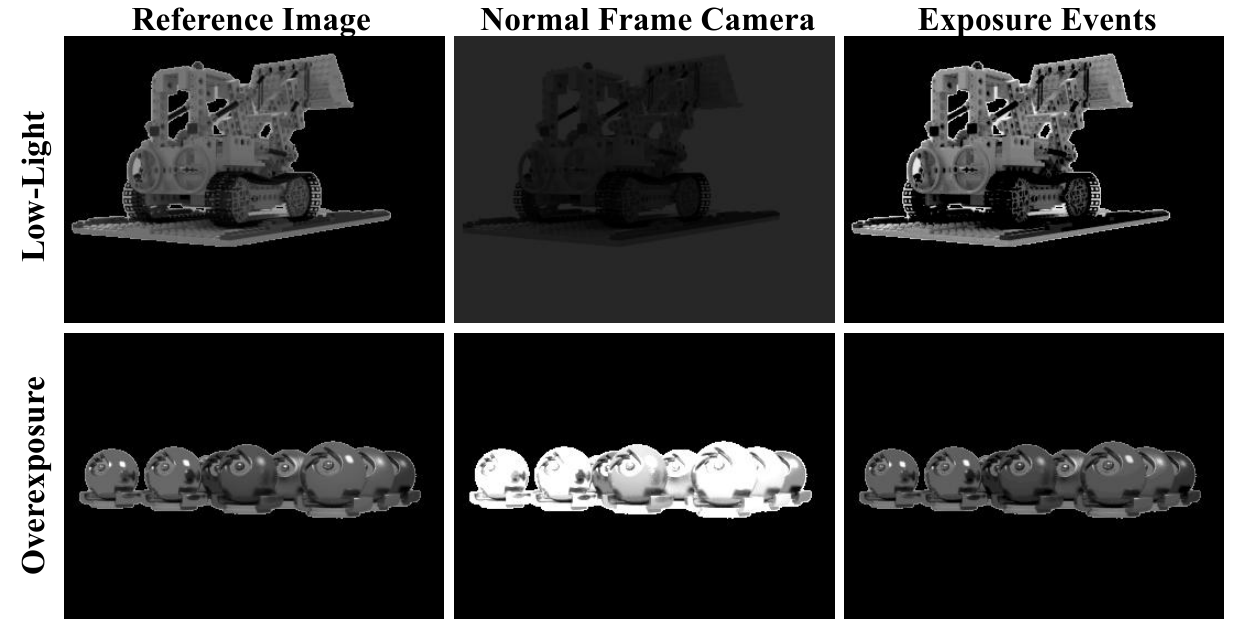}
% \vskip -3ex
\caption{Comparison of standard frame-based cameras and grayscale images generated from exposure events under extreme lighting conditions. The first column shows grayscale images derived from EventNeRF~\cite{rudnev2023eventnerf} ground truth RGB images with foreground masks extracted using EfficientSAM~\cite{xiong2023efficientsam}, serving as references under favorable lighting conditions. The second column presents simulated frame-based images under low light and overexposure. The third column presents grayscale images mapped from exposure events, derived from the frame-based images using the method described in Section~\ref{Experiments:Evaluation Dataset}. These images exhibit improved performance in low-light conditions and demonstrate the capability to recover details lost in overexposed regions, leveraging the high dynamic range and temporal resolution of event cameras.
}
\label{fig:EvTemMap_RGB_compare}
\end{figure*}

\subsection{Qualitative Comparison}
We qualitatively compare the appearance and geometric reconstruction of five sequences from the real-world EME-3D dataset by visualizing the rendered results and depth maps. 
Fig.~\ref{fig:render_compare} shows that our E-3DGS, in both fast and high-quality reconstruction modes, preserves sharper, more consistent structures and cleaner backgrounds compared to EventNeRF~\cite{rudnev2023eventnerf}. When using motion events, E-3DGS avoids artifacts seen in EventNeRF’s rendering, such as with the car ($1^{st}$ row) and the tortoise ($4^{th}$ row). Our results also exhibit higher contrast and sharper spatial details, particularly in regions with highlights and reflections, like the camera ($3^{rd}$ row) and the tortoise ($4^{th}$ row).
Compared to E2VID~\cite{Rebecq19cvpr} + NeRF, E-3DGS produces clearer local details, such as the tortoise’s shell texture, which is completely lost in the $4^{th}$ row of E2VID + NeRF. 
Furthermore, compared to E2VID + 3DGS, E-3DGS integrates exposure event information and reconstructs the overall grayscale tone and brightness more accurately. 
For instance, in the $3^{rd}$ row, E2VID + 3DGS shows white artifacts on the black camera body, while E-3DGS renders a clean, pure black camera body.

As shown in Fig.~\ref{fig:depth_compare}, E-3DGS achieves cleaner backgrounds, fewer artifacts, and richer spatial details in geometric reconstruction compared to both EventNeRF and NeRF integrated with E2VID. 
EventNeRF's depth maps struggle to distinguish foreground from background and suffer from noticeable noise, such as with the car ($1^{st}$ row), motorcycle ($2^{nd}$ row), tortoise ($4^{th}$ row), and crab ($5^{th}$ row). 
While the depth map quality of E2VID + 3DGS is comparable to E-3DGS, E-3DGS excels in high-frequency spatial details. 
For example, in the crab sample, E-3DGS successfully reconstructs the gap between the crab's body and the base, while E2VID + 3DGS confuses the subject with the background.
Furthermore, the depth maps obtained in the fast reconstruction mode exhibit contours that align more closely with the actual geometry of the objects, providing a significant advantage when converting the 3DGS representation into mesh structures.

Previous qualitative and quantitative experiments have demonstrated the photometric accuracy advantage of our High-Quality Reconstruction Mode in utilizing exposure events for 3D reconstruction compared to motion events. 
While capturing exposure events may seem to compromise the event camera's ability to handle fast-moving scenes, it retains its inherent advantage over standard frame-based cameras under extreme lighting conditions. 
As shown in Fig.~\ref{fig:EvTemMap_RGB_compare}, we simulate grayscale images mapped from exposure events under low-light and overexposure conditions using the method described in Section~\ref{Experiments:Evaluation Dataset}. 
Under low-light conditions, standard frame-based cameras lose the ability to effectively capture contour and fine structural details due to their limited dynamic range and reduced signal-to-noise ratio. 
In contrast, grayscale images mapped from exposure events exhibit a significantly higher dynamic range, enabling better detail preservation. 
Furthermore, in cases of overexposure, RGB cameras struggle with pixel saturation, where the sensor reaches its maximum capacity to record light intensity. 
This saturation leads to a loss of crucial details in high-brightness regions, such as reflective surfaces or intense light sources, making these areas appear as uniform white with no discernible features.  
However, since event cameras record brightness changes rather than absolute intensity, the grayscale images mapped from exposure events can effectively retain details in overexposed areas, showcasing their robustness in extreme illumination scenarios.

\subsection{Ablations}
In the ablation study, unless otherwise specified, we train on each sequence of the real-world EME-3D dataset and report the average accuracy across all sequences. We investigate various aspects, including point cloud initialization methods for event-based 3DGS, the accuracy variation in the balanced hybrid mode of E-3DGS, and the accuracy difference between novel view synthesis and 3D reconstruction tasks using E-3DGS.

\begin{figure}[!t]
\centering
\includegraphics[width=0.8\linewidth]{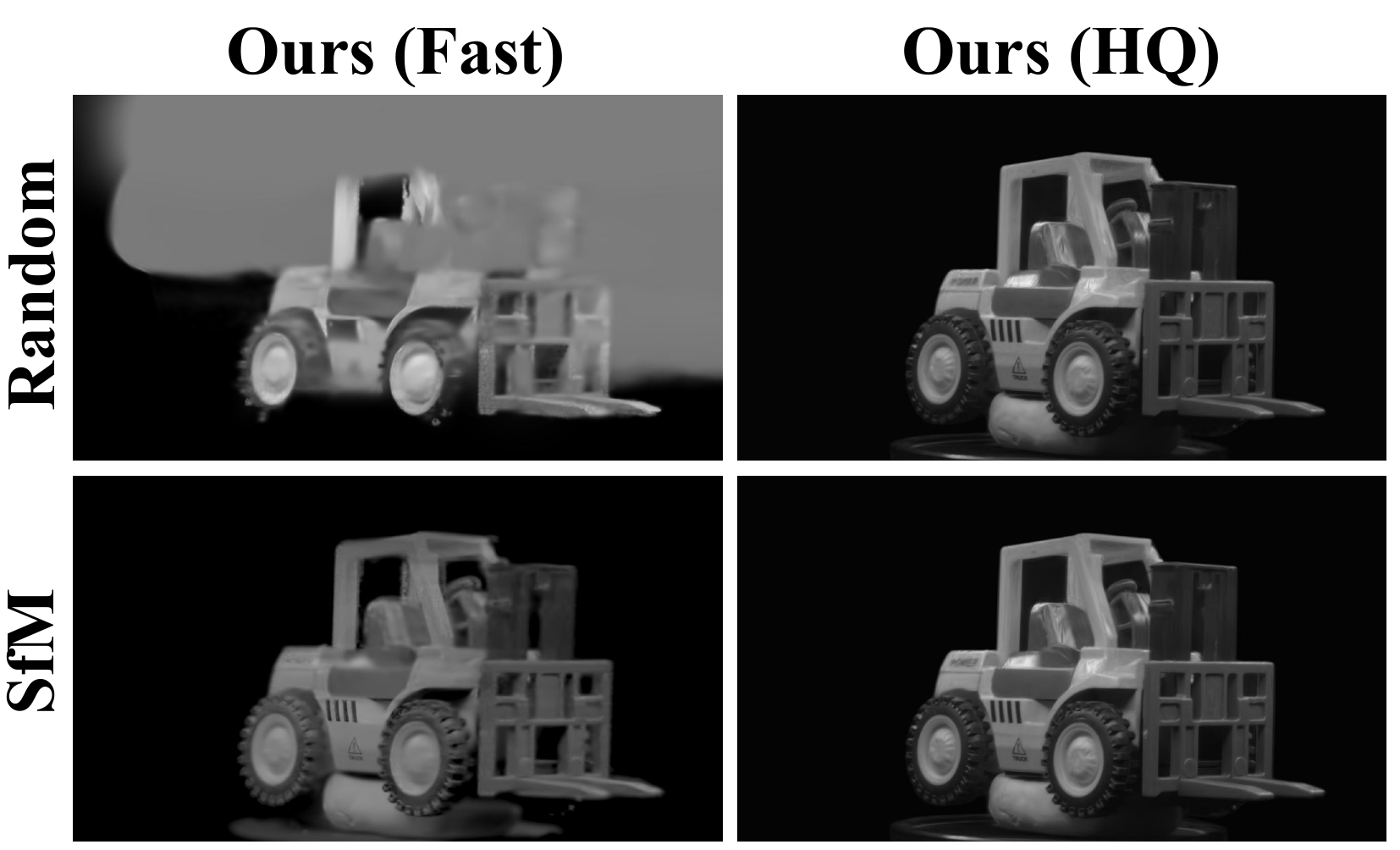}
\caption{Qualitative comparison of different initialization methods. The results demonstrate the impact of initialization on the performance of motion event-based and exposure-enhanced 3DGS approaches. When using random initialization, the motion event-based 3DGS method produces noticeable artifacts and lacks fine details, as observed in the top-left image compared to the bottom-left. In contrast, the E-3DGS method, which integrates spatially rich exposure events, shows significantly improved robustness to initialization. As seen in the top-right and bottom-right images, the rendering results remain consistent and free of significant artifacts, regardless of the initialization method.
}
\label{fig:compare_initialization}
\end{figure}

\begin{table}[!t]
    \centering
    \caption{Impact of Initialization Methods on Fast and HQ Modes.}
    \begin{adjustbox}{width=0.6\columnwidth}
    \begin{tabular}{llcccc}
        \toprule
        \textbf{Metric} & \textbf{Method} & \textbf{Random} & \textbf{SfM~\cite{snavely2006photo}} & \textbf{Difference} \\
        \midrule
        \multirow{2}{*}{\textbf{PSNR (dB)}} 
        & Ours (Fast) & 15.72 & 25.22 & + 60 $\%$ \\
        & Ours (HQ) & 33.83 & 33.92 & + 0.3$\%$ \\
        \midrule
        \multirow{2}{*}{\textbf{SSIM} $\uparrow$} 
        & Ours (Fast) & 0.21 & 0.89 & + 324 $\%$ \\
        & Ours (HQ) & 0.96 & 0.97 & + 1.04 $\%$ \\
        \midrule
        \multirow{2}{*}{\textbf{LPIPS} $\downarrow$} 
        & Ours (Fast) & 0.58 & 0.14 & 75.9 $\%$ \\
        & Ours (HQ) & 0.09 & 0.09 & - \\
        \bottomrule
    \end{tabular}
    \end{adjustbox}
    \label{tab:Ablation_initialization}
\end{table}

\begin{table}[!t]
    \centering
    \caption{Ablation Study on Different Initialization Methods (Fast Mode).}
    \begin{tabular}{lccc}
    \toprule
    \textbf{Method} & \textbf{PSNR $\uparrow$} & \textbf{SSIM $\uparrow$} & \textbf{LPIPS $\downarrow$} \\
    \midrule
    Random Initialization & 15.72 & 0.21 & 0.58 \\
    EventNeRF~\cite{rudnev2023eventnerf} & 17.13 & 0.72 & 0.29 \\
    E2VID+SfM~\cite{snavely2006photo} & 23.47 & 0.87 & 0.15 \\
    \textbf{Exposure+SfM~\cite{snavely2006photo}} & \textbf{25.22} & \textbf{0.89} & \textbf{0.14} \\
    \bottomrule
    \end{tabular}
    \label{tab:Ablation_initialization_method}
\end{table}

\textbf{Point Cloud Initialization.} 
In Tab.~\ref{tab:Ablation_initialization}, we explore the impact of different point cloud initialization methods on rendering performance in event-based 3DGS. 
Compared to random initialization, using a sparse point cloud from SfM~\cite{snavely2006photo} significantly improves the rendering accuracy when using motion events, with PSNR elevating to $25.22dB$ from $15.72dB$. 
However, in high-quality reconstruction mode, the PSNR is not sensitive to whether the initialization is from SfM or random point clouds ($33.82dB$ \textit{vs.} $33.92dB$).
As shown in Fig.~\ref{fig:compare_initialization}, we visualize the effects of different initialization methods on event-based 3DGS rendering. When using random initialization, motion event-based 3DGS produces noticeable artifacts and limited details (top-left \textit{vs.} bottom-left). In contrast, after integrating the spatially rich exposure events, E-3DGS becomes more robust to the initialization method, with no significant differences in rendering results (top-right \textit{vs.} bottom-right).
To further illustrate the advantages of incorporating exposure events in 3D reconstruction, we compare NeRF-based and event-to-video-based point cloud initialization methods in Tab.~\ref{tab:Ablation_initialization_method}. Initializing the point cloud using EventNeRF-rendered results, compared to random initialization, provides a modest improvement in PSNR ($17.13dB$ \textit{vs.} $15.72dB$), though it requires additional training for EventNeRF with limited gains. Using E2VID~\cite{Rebecq19cvpr} to convert events into images followed by SfM for point cloud initialization further enhances accuracy ($23.47dB$ \textit{vs.} $15.72dB$), but this process also involves additional learning-based computation.
By fully leveraging the rich spatial cues embedded in exposure events, the proposed exposure event temporal-spatial mapping combined with SfM initialization yields the greatest improvement in reconstruction quality ($25.22dB$ \textit{vs.} $15.72dB$). 
This certifies the critical role of exposure events in enhancing the quality of explicit 3D reconstruction from events.

\begin{figure}[!t]
\centering
\includegraphics[width=0.8\linewidth]{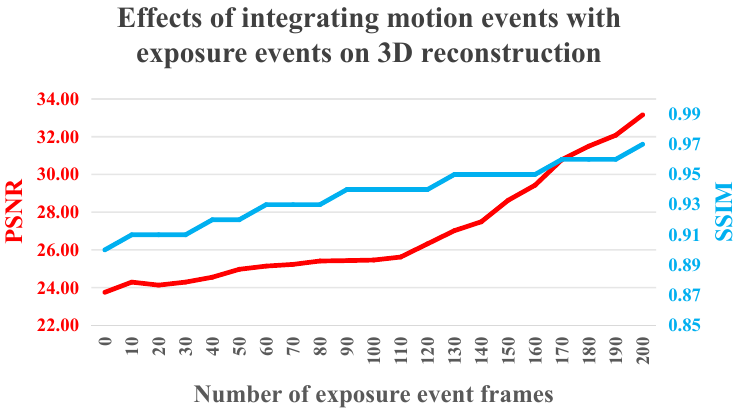}
\caption{Effects of integrating motion events with exposure events on 3D reconstruction for the real-world car sequence. Incorporating exposure event frames provides more accurate spatial information to constrain the 3D Gaussian splatting optimization, leading to steady improvements in PSNR and SSIM. Even a small number of exposure frames enhances reconstruction quality compared to using only motion events, highlighting the balanced hybrid mode's effectiveness in fast-motion scenarios.}
\label{fig:Ex_Mo}
\end{figure}

\begin{table}[!t]
    \centering
    \caption{Quantitative Results of Novel View Reconstruction.}
    \begin{adjustbox}{width=0.75\columnwidth}
    {
    \begin{tabular}{lcccccccc}
    \toprule
    Sequence & \multicolumn{4}{c}{Ours (Fast Reconstruction)} & \multicolumn{4}{c}{Ours (High-Quality Reconstruction)} \\
    \cmidrule(lr){2-5} \cmidrule(lr){6-9}
    & \multicolumn{2}{c}{PSNR} & \multicolumn{2}{c}{SSIM} & \multicolumn{2}{c}{PSNR} & \multicolumn{2}{c}{SSIM} \\
    \cmidrule(lr){2-3} \cmidrule(lr){4-5} \cmidrule(lr){6-7} \cmidrule(lr){8-9}
    & Given & Novel & Given & Novel & Given & Novel & Given & Novel \\
    \midrule
    car & 22.98 & 22.96 & 0.88 & 0.87 & 25.60 & 25.42 & 0.91 & 0.91 \\
    motorcycle & 25.22 & 24.94 & 0.90 & 0.89 & 27.34 & 27.06 & 0.92 & 0.92 \\
    camera & 23.99 & 24.05 & 0.87 & 0.87 & 26.42 & 26.43 & 0.90 & 0.90 \\
    cat & 25.35 & 25.34 & 0.91 & 0.90 & 26.75 & 27.34 & 0.93 & 0.92 \\
    crab & 22.21 & 22.19 & 0.87 & 0.87 & 23.11 & 23.05 & 0.89 & 0.89 \\
    forklift & 19.92 & 19.90 & 0.76 & 0.76 & 21.37 & 21.23 & 0.81 & 0.81 \\
    racing car & 23.83 & 23.74 & 0.88 & 0.88 & 25.73 & 25.57 & 0.91 & 0.91 \\
    tortoise & 22.55 & 22.44 & 0.88 & 0.88 & 24.26 & 24.10 & 0.91 & 0.91 \\
    train & 21.80 & 21.77 & 0.82 & 0.82 & 23.41 & 23.30 & 0.85 & 0.85 \\
    Average & 23.09 & 23.04 & 0.86 & 0.86 & 24.89 & 24.83 & 0.89 & 0.89 \\
    \bottomrule
    \end{tabular}
    }
    \end{adjustbox}
  \label{tab:novel_view}
\end{table}

\textbf{Balanced Hybrid Mode.} 
We further look into how to balance reconstruction quality and the number of exposure event frames in E-3DGS. 
To handle fast and complex scenes, E-3DGS includes a balanced hybrid mode. 
In this mode, we control the aperture to capture a small number of exposure events during the initial phase of a sample, followed by collecting only motion events to achieve high-quality reconstruction of fast-moving scenes.
In Fig.~\ref{fig:Ex_Mo}, we show the impact of the number of exposure event frames on rendering quality. As the number of exposure event frames increases, more accurate spatial information is used to constrain the optimization of the Gaussian splatting, leading to a steady improvement in PSNR and SSIM of the rendered images. In other words, introducing even a small number of exposure event frames provides a positive gain in 3D reconstruction quality compared to using none. This highlights the effectiveness and potential of the balanced hybrid mode for handling fast-motion scenes.

\textbf{Novel View Synthesis.} 
In this task, we train the model using the odd-numbered frames (Given) and test its ability to generate novel views using the even-numbered frames (Novel).
As shown in Tab.~\ref{tab:novel_view}, E-3DGS has excellent novel view synthesis performance in both fast reconstruction mode and high-quality reconstruction mode. 
On average, E-3DGS using motion events exhibits only a $0.05dB$ drop in PSNR for novel view synthesis compared to the given frames ($23.04dB$ \textit{vs.} $23.09dB$). 
Similarly, E-3DGS using exposure events shows a $0.06dB$ PSNR drop for novel views compared to the given frames ($24.83dB$ \textit{vs.} $24.89dB$). 
This demonstrates E-3DGS's strong capabilities in both 3D reconstruction and novel view synthesis tasks.

\section{Conclusion and Discussions}
\textbf{Conclusion.}
In this work, we have presented E-3DGS, the first method to integrate 3DGS with exposure events for explicit scene reconstruction using a single pure high-resolution event sensor. 
By partitioning events into motion and exposure categories, E-3DGS leverages motion events for efficient 3D Gaussian Splatting (3DGS) reconstruction and utilizes exposure events to enhance reconstruction quality through high-resolution temporal-to-intensity mapping. 
Our framework is versatile and capable of operating on motion/exposure events alone or adopting a hybrid mode that balances reconstruction quality and speed by combining initial exposure events with high-speed motion events. 
On the EventNeRF dataset, we validate the feasibility of integrating a transmittance adjustment device with event cameras for high-quality 3D reconstruction. 
Under extreme lighting conditions, such as low illumination and overexposure, conventional frame-based cameras struggle with detail loss due to limited dynamic range. 
In contrast, the hardware-level adjustment preserves the high dynamic range advantage of event cameras, making them more effective for imaging in challenging environments.
Additionally, we introduced EME-3D, a real-world dataset with exposure and motion events, camera calibration parameters, and sparse point clouds to support further research in this domain.
E-3DGS outperforms event-based NeRF in both reconstruction speed and quality while being more cost-effective than methods that require both event and RGB data. 
By fully utilizing both motion and exposure events, E-3DGS sets a new benchmark for event-based 3D reconstruction, demonstrating robust performance in challenging conditions with reduced hardware demands.

\textbf{Limitations.}
Despite the promising results, our current work has several limitations. 
Firstly, our method currently reconstructs grayscale scenes, lacking color information inherent in traditional RGB-based approaches. This limits its applicability in domains where color is essential for interpretation or downstream tasks. Secondly, we have not evaluated performance under extremely high event rates, where event camera saturation could degrade results, particularly in scenarios like large scenes under bright sunlight. Our system currently lacks specific mechanisms for handling such conditions. Thirdly, experiments were confined to smaller datasets, and the method's scalability to large-scale environments remains untested.

\textbf{Future work.}
Future work should prioritize addressing these limitations. Key directions include developing strategies for robustness against high event rates, such as adaptive event processing or integrating saturation models. Extending and evaluating E-3DGS on large-scale scenes is another critical goal, likely requiring algorithmic optimizations for efficiency and enhanced robustness to diverse real-world conditions.
Crucially, we plan to explore the integration of color information, potentially by leveraging color event cameras and developing algorithms capable of reconstructing colored 3D scenes from their sparse, asynchronous output. This would significantly broaden the applicability of our event-based reconstruction framework.
Furthermore, we intend to investigate advanced fusion algorithms and hardware optimizations to further exploit the complementary strengths of both exposure events and motion events, potentially leading to hybrid systems effective across an even wider range of conditions.

\noindent \textbf{Funding.} Natural Science Foundation of Zhejiang Province (Grant No. LZ24F050003); National Natural Science Foundation of China (Grant Nos. 12174341 and 62473139). 

\noindent \textbf{Disclosures.} The authors declare that there are no conflicts of interest related to this article.

\noindent \textbf{Data availability.} The data underlying the results presented in this paper will be publicly available on GitHub at \url{https://github.com/MasterHow/E-3DGS}.

%%%%%%%%%%%%%%%%%%%%%%% References %%%%%%%%%%%%%%%%%%%%%%%%%

\bibliography{sample}

\end{document}